\documentclass[10pt,journal]{article}

\usepackage[english]{babel}
\usepackage[T1]{fontenc}
\usepackage{amssymb}
\usepackage{amsfonts}
\usepackage{amsmath}
\usepackage{amsthm}
\usepackage{mathpazo}
\usepackage{bm}
\usepackage[colorinlistoftodos]{todonotes}
\usepackage{soul}
\usepackage{enumitem}
\usepackage{float}
\usepackage{rotating}
\usepackage{multirow}
\usepackage[utf8]{inputenc} 
\usepackage{algorithm}
\usepackage{algpseudocode}
\usepackage[colorlinks=true, allcolors=blue]{hyperref}
\usepackage{lipsum,graphicx}
\usepackage{hhline}
\usepackage{setspace}
\usepackage[top=2.54cm,bottom=2.54cm, left=2.54cm, right=2.54cm]{geometry}
\usepackage{epstopdf}
\usepackage{xcolor}

\DeclareMathOperator*{\argmax}{arg\,max}

\title{A generalised OMP algorithm for feature selection with application to gene expression data}
\author{Michail Tsagris$^1$, Zacharias~Papadovasilakis$^2$, Kleanthi~Lakiotaki$^3$ and
Ioannis~Tsamardinos$^4$ \\
\\
$^1$ Department of Economics, University of Crete, Rethymnon, Greece,  \\
\href{mailto:mtsagris@uoc.gr}{mtsagris@uoc.gr} \\
$^2$ Department of Computer Science, University of Crete, Herakleion, Greece, \\ \href{mailto:zpapadov@hotmail.com}{zpapadov@hotmail.com} \\
$^3$ Department of Computer Science, University of Crete, Herakleion, Greece, \\ \href{mailto:kliolak@gmail.com}{kliolak@gmail.com} \\
$^4$ Department of Computer Science, University of Crete and Institute of Applied and \\
Computational Mathematics, FORTH, Herakleion, Greece, \\ 
\href{mailto:tsamard.it@gmail.com}{tsamard.it@gmail.com}
}

\begin{document}

\maketitle

\begin{center}
{\bf Abstract}
\end{center}
Feature selection for predictive analytics is the problem of identifying a minimal-size subset of features that is maximally predictive of an outcome of interest. To apply to  molecular data, feature selection algorithms need to be scalable to tens of thousands of available features. In this paper, we propose gOMP, a highly-scalable generalisation of the Orthogonal Matching Pursuit feature selection algorithm to several directions: (a) different types of outcomes, such as continuous, binary, nominal, and time-to-event, (b) different types of predictive models (e.g., linear least squares, logistic regression), (c) different types of predictive features (continuous, categorical), and (d) different, statistical-based stopping criteria. We compare the proposed algorithm against LASSO, a prototypical, widely used algorithm for high-dimensional data. On dozens of simulated datasets, as well as, real gene expression datasets, gOMP is on par, or outperforms LASSO for case-control binary classification, quantified outcomes (regression), and (censored) survival times (time-to-event) analysis. gOMP has also several theoretical advantages that are discussed. While gOMP is based on quite simple and basic statistical ideas, easy to implement and to generalize, we also show in an extensive evaluation that it is also quite effective in bioinformatics analysis settings. 
\\
\\
\textbf{Keywords}: 
Feature selection, high dimensional data, bioinformatics, omics data, gene expression data.


\section{Introduction}\label{sec:introduction}
The problem of feature selection (hereafter FS, a.k.a. variable selection), in a predictive analytics context, is the problem of identifying a minimal-size, optimally predictive subset $\mathcal{S}$ out of available features $\mathcal{X}$ for an outcome $\mathcal{Y}$ (calligraphic letters denote random variables). To solve the problem, we assume we are given a dataset $X$ where the $n$ rows correspond to samples and the columns correspond to features, such that each row is vector $x$ of values of $\mathcal{X}$, and the vector $Y$ of the corresponding values of $\mathcal{Y}$ for each sample.

FS is commonly employed in predictive and diagnostic analytics for several reasons. It often improves the predictive performance of the resulting model: removing irrelevant or redundant features facilitates the task of the model-learning algorithm. This is especially true for algorithms susceptible to the curse of dimensionality. It also results in models that are faster to execute, and easier to visualize, inspect, understand, and interpret. In addition, FS can be used as a means of knowledge discovery and for gaining intuition on the data generation mechanisms. Indeed, there is a theoretical connection between the solution to the FS problem (i.e., the optimal selected feature subset) and the generative Bayesian (causal) network \cite{tsamardinos2003towards}.

For these reasons, FS is commonly employed in high-dimensional, molecular data and is of paramount importance in bioinformatics. In fact, it is often the case that FS is the {\em primary} goal of the analysis, while the predictive model is a by-product. For example, a biologist may be interested in which genes are diagnostic of a disease as a means to gain intuition into the biological mechanism. A bioengineer may perform FS to identify a minimal-size feature subset to design a cost-effective diagnostic assay. A pharmacologist may use FS to identify potential drug targets. 

FS algorithms for use in bioinformatics are required to scale up to tens or even hundreds of thousands of features (e.g., multi-omics data) and maintain high-quality, even when sample sizes are relatively small (e.g., a few hundreds). In addition, they have to be general enough for binary outcomes (e.g., classifying between disease cases and controls), multi-class outcomes (e.g., classifying between disease subtypes), continuous outcomes (e.g., quantitative traits), and censored time-to-event outcomes (e.g,. time to death, complications, relapse, metastasis). In addition, while features may often be only continuous (e.g., gene expressions), a FS algorithm should allow for categorical features too to be able to also consider clinical and single nucleotide polymorphism data.

The FS literature is vast (some recent reviews in \cite{saeys2007,buhlmann2011,bolon2014,bolon2016,ang2016}); in this paper however, we focus only on the most prominent algorithms that can abide to the requirements above, namely, scalability to the number of features and ability to generalize to several types of outcomes and features. Most such high-dimensional algorithms are greedy in their selection (see \cite{bertsimas2016} for exceptions). Specifically, we focus and generalize the Orthogonal Matching Pursuit (OMP) algorithm, a quite popular algorithm in the signal processing literature \cite{pati1993,mallat1993,davis1994}, and generalize it to become an FS algorithm named generalized OMP or gOMP. 

The main intuition of gOMP is to insert next in the selected feature subset $\mathcal{S}$ the feature that is mostly pair-wise associated the residuals (errors) of the current model, until a stopping criterion is satisfied. These ideas are then instantiated for different problem settings in computationally efficient and statistically robust ways. Residuals and pair-wise associations can be defined and computed for numerous outcome and feature types. In addition, for typical pair-wise associations (e.g,. Pearson linear correlation and Spearman correlation for continuous features or t-test for a binary feature) that can be computed efficiently, the algorithm scales up to millions of features in reasonable time. When different types of features are considered (e.g,. continuous and categorical), the solution is to convert associations to $p$-values so they are in the same scale and directly comparable. Any model for the given outcome type can serve as the statistical model on which to compute residuals, e.g., (robust or least-squares) linear regression for continuous outcomes and Cox proportional hazards for time-to-event. Since a new model is trained only for every newly selected feature, gOMP is relatively efficient in cases where the total number of selected features is relatively small (< 100), a typical situation in bioinformatics. A last generalization regards the stopping criterion. The original OMP stops selection when it reaches a low signal reconstruction error, i.e., low mean squared error. In a FS context, we define statistical-based stopping criteria for gOMP. For example, we consider stopping when the Bayesian Information Criterion (BIC) is not sufficiently improved for the next feature selected. Such criteria can be computed for most statistical-based predictive models that estimate the data likelihood. 

We contrast gOMP to LASSO \cite{tibshirani1996}, arguably the most popular FS algorithm in bioinformatics, machine learning and statistics. LASSO is also computationally scalable to the typical feature sizes in bioinformatics, exhibits excellent predictive quality, and has been generalized to different types of outcomes (multivariate, binary, multinomial, ordinal, time-to-event, clustered) and categorical features \cite{yang2017}. In addition, LASSO is also quite similar in spirit to gOMP: while LASSO solves a global optimization problem, it can be proven that it can be solved in a greedy fashion, similar to gOMP \cite{efron2004}. The results of the comparative evaluation show that \textit{gOMP is on par, or outperforms LASSO in several aspects}. Specifically, our results are summarised as follows:
\begin{enumerate}
\item In simulated data where the FS solution is known, gOMP consistently exhibits fewer false positive selections. It also outperforms LASSO in terms of predictive performance, for the same number of selected features. 
\item In real gene expression data, gOMP produces predictive models of similar performance, while selecting fewer features than LASSO. When contrasted to LASSO on an equal footing (roughly equal number of selected features), we draw conclusions similar to the simulation studies.
\item gOMP is computationally more efficient than LASSO and scales up to higher number of  dimensions and/or larger sample sizes better than LASSO. 
\end{enumerate}
gOMP is based on simple ideas and a particularly greedy heuristic for selecting the next best feature. Yet, in this evaluation we show that it is not only scalable and general, but also quite competitive in term of predictive performance. gOMP is easy to implement and to generalize to other settings; in contrast, LASSO's optimization problem is not convex for some outcomes or predictive models and requires task-specific algorithms to generalize to other settings. For example, for time-course data LASSO is non-convex and non-scalable.

\section{A Unifying View of Greedy Feature Selection algorithms}

Solving the FS problem is inherently a combinatorial problem that is worst-case NP-complete, even for linear models \cite{borboudakis2019}. Hence, for high-dimensional data, most algorithms rely on some sort of greedy strategy to include the next feature to select in $\mathcal{S}$, or to remove a feature from $\mathcal{S}$. A particular class of algorithms maintain a current predictive model $M_{\mathcal{S}}$ using only the selected features $\mathcal{S}$. It then inserts to $\mathcal{S}$, the feature $X'$ that improves $M$ the most, reaching $M_{\mathcal{S}\cup X'}$. We will call them {\em model-based} in this context. Some employ a similar idea to drop next the feature that deteriorates  
$M_{\mathcal{S}\setminus X'}$
the least. Algorithms may include both insertion and deletion, either in separate phases or as interleaved steps. Examples include Forward Search, Forward-Backward Search, Orthogonal Least Squares, Incremental Association Markov Blanket, the Grow-Shrink algorithm, and the recent Forward-Backward with Early Dropping (FBED) algorithm. The two main unifying ideas of the model-based family are that (a) they maintain a model fit with all selected features, and (b) to identify the best next feature, they train new models for each feature under consideration. 

A second family of algorithms differs in the greedy heuristic employed: they select as the next feature the one that maximizes the pairwise association (correlation in a general sense) with the residuals of the current model. We will call them {\em residual-based}. Examples of residual-based algorithms are Orthogonal Matching Pursuit (OMP), Least Angle Regression (LARS), and Forward Step-wise Regression (FSR). LASSO is an algorithm that is expressed as a global optimization problem, however, it is still solvable using a residual-based greedy search. A details discussion of these algorithms and families with references to each individual algorithm can be found in \cite{borboudakis2019}.

The residual-based family is expected to be computationally more efficient as pairwise associations are faster to estimate than training a new model for each feature under consideration. This statement is not absolute of course. Computing non-linear pairwise associations may become quite expensive \cite{}, while training new models for discrete data may be as simple as creating contingency tables. The question of course, arises regarding the comparative quality of the residual-based algorithms. 

We'd like to note the presence of another family of algorithms that differs in the sense that it does not maintain a single model $M_{\mathcal{S}}$ trained with all selected features. Instead, it computes its heuristic values to select features based on subsets of $\mathcal{S}$ of smaller size. Let us call this family {\em reduced-model} algorithms. Examples include the HITON algorithm, the Max Min Parents and Children (MMPC), and the Statistically Equivalent Signatures (SES) \cite{tsamardinos2012, lagani2017}. Similarly, all information theoretic algorithms belong in this category, as they compute mutual information considering only subsets of the current $\mathcal{S}$ \cite{brown2012}. 

\section{OMP and LASSO algorithms}
We now present the two algorithms mostly related to the current work, namely OMP which forms the basis for the proposed gOMP, and LASSO against which we focus the evaluation.

\makeatletter
\def\BState{\State\hskip-\ALG@thistlm}
\makeatother

\begin{algorithm}[H]
\label{alg:omp} 
\centering
\begin{algorithmic}[1]
\BState \textbf{Input}: \text{Outcome values $y$, dataset $X$} 
\State \textbf{Output}: \text{A subset $\mathcal{S}\subseteq \mathcal{X}$ of selected features.} 
\State Standardize the data: Center each feature and $y$ and scale it to unit norm 
\State // Generalize residuals to other types of outcomes
\State Initialize residuals ${r}\leftarrow {y}$
\State $\mathcal{S}\leftarrow \emptyset$
\State // Generalize to statistical-based criteria
\While {$|| {\bf r} ||^2 > \epsilon $}
\State // Generalize to other types of association
\State // Generalize to other types of features
\State // Convert associations to p-values to compare
\State $X_* \leftarrow \argmax_{i \in \mathcal{X} \setminus \mathcal{S}}| \langle {r}, {X}_i \rangle |$ 
\State $\mathcal{S} \leftarrow \mathcal{S}\cup \{X_*\}$
\State // Generalize to other types of 
\State // statistical predictive models
\State $\beta \leftarrow$ least squares regression of $y$ on $X_\mathcal{S}$
\State // Update residuals
\State $r \leftarrow y - X_{\mathcal{S}}\cdot{\beta}$
\EndWhile \\
\Return $\mathcal{S}$
\end{algorithmic}
\caption{The OMP algorithm} \label{omp}
\end{algorithm}

Orthogonal Matching Pursuit (OMP) (\cite{chen1989,pati1993,davis1994}) is a greedy forward search algorithm that was first proposed for continuous outcomes in the context of signal reconstruction. 

{\em OMP assumes continuous outcome ($y$) and a data matrix $X$ of continuous features} and its pseudo-code is in Algorithm \ref{omp}. The $i$th column of $X$ is denoted as $X_i$, while $X_\mathcal{S}$ denote the matrix with the columns of all selected features. The algorithm initiates the current selection $\mathcal{S} \leftarrow \emptyset$, and residuals $r \leftarrow y$. At each iteration, the feature with the largest Pearson linear correlation with $r$ is selected for inclusion. If both $X$ and $r$ are centered and normalized, then correlation of $r$ with $X_i$ is simply $\langle r, X_i\rangle$, i.e., the inner product of the two vectors. After selection, a new least squares regression model is fit resulting in coefficients $\beta$; it is used to update the residuals. The least squares solution results in centered residuals. The procedure stops when the 2-norm of the residuals is below some threshold $\epsilon$ (alternatively, the mean squared error, or the mean absolute error can be used). In signal reconstruction applications, the signal corresponds to the outcome $y$. It can be perfectly reconstructed using all features, hence the residuals can go arbitrarily go to zero. For FS applications however, the criterion may never be satisfied, even when all features are selected. The comments in Algorithm \ref{alg:omp} indicate the generalizations of gOMP that are presented next. 

OMP is similar to the residual-based FSR (\cite{weisberg1980, efron2004}). The main difference is that the latter selects a feature if its correlation with the current residual is statistically significant, or if its magnitude is above a certain threshold \cite{stodden2006}. 

\subsection{LASSO algorithm}
Least Absolute Shrinkage and Selection Operation (LASSO) \cite{tibshirani1996} is perhaps the most popular FS algorithm, owing much of its success to its computational efficiency, and high learning quality with both low and medium sample sizes. LASSO is formulated around the penalized minimization (regularization) of an objective function (usually log-likelihood), which depends upon the type of the outcome variable. The level of penalization affects the magnitude of the regression coefficients forcing them to shrink towards zero, hence regularization of the coefficient values and feature selection are performed simultaneously. The two main ideas of LASSO are that the predictive model employed is a generalized linear model that is expressed with a set of coefficients $\beta$ and that the regularization penalty is the $l_1$ norm of the coefficient vector (the sum of absolute values of the $\beta$'s). In contrast, ridge regression is penalizing using the $l_2$ norm of $\beta$. Ridge regression has an analytic solution for linear regression, but does not perform feature selection, leaving most coefficients to non-zeros. LASSO on the other hand, requires iterative optimization techniques to solve, but zero's more coefficients, implicitly performing feature selection and modeling at the same time.

LASSO for continuous outcomes was first solved effectively in 2004 by Efron et al. \cite{efron2004}) using a residual\footnote{This so called LARS solution used the raw residuals $y-\hat{y}$}-based forward selection method. The procedure is a modified version of Least Angle Regression (LARS) borrowing ideas from OMP. A main difference with OMP is that LARS takes into account the constraint on the regression coefficients. In 2010 though, Friedman et al. \cite{friedman2010} proposed a coordinate descent method that is applicable to a wider range of outcomes.

For binary outcomes ($\left\lbrace0, 1\right\rbrace$) modelled via logistic regression, the objective function to be minimised over $\beta$ is:
\begin{eqnarray}  \label{loglasso}
\sum_{i=1}^n\left[y_i \sum_{j=1}^p\beta_jx_{ij} - \log{\left(1+e^{\sum_{j=1}^p\beta_jx_{ij}} \right)} \right] + \lambda \sum_{j=1}^p|\beta_j|,
\end{eqnarray}
whereas with continuous outcomes and linear regression the objective function is:
\begin{eqnarray}  \label{linlasso}
\sum_{i=1}^n\left(y_i - \sum_{j=1}^p\beta_jx_{ij}\right)^2 + \lambda \sum_{j=1}^p|\beta_j|.
\end{eqnarray}
With time-to-event (strictly positive continuous) data, using the partial log-likelihood of the Cox regression we get \cite{friedman2010}
\begin{eqnarray} \label{coxlasso}
- \sum_{i=1}^n{\bf x}_{j(i)}^T\pmb{\beta} + \sum_{i=1}^n\log{\left(\sum_{j \in R_i}e^{{\bf x}_j^T\pmb{\beta}} \right)} + \lambda \sum_{j=1}^p|\beta_j|,
\end{eqnarray}
where $R_i$ is the set of indices $j$ with $y_j \geq t_i$ (those at risk at time $t_i$) and $\pmb{\beta} = \left(\beta_1, \ldots, \beta_p\right)$ denote the vector of regression coefficients. In all cases, the $\lambda$ is a positive valued hyper-parameter that must be tuned (see Section \ref{tuning}). 

\subsection{OMP vs. LASSO} 
Both OMP and LASSO can be thought of as residual-based algorithms, i.e. they select the next best feature based on the residuals produced by the regression model. Unlike OMP, LASSO may also drop features from the selected set $\mathcal{S}$. While LASSO's penalty uses the $l_1$ norm of $\beta$, OMP is conceptually trying to solve the same objective function (minimize loss) where the penalty is now the zero-norm ($l_0$) of the coefficients, i.e., {\em the number} of non-zero beta's. The optimization is performed in a greedy fashion without guarantees of optimalities.

The computational complexity of LASSO, for linear models, is $\mathcal{O}(np^2)$ \cite{rosset2007}, where $n$ and $p$ denote the sample size and number of features, respectively. OMP on the other hand performs $p \cdot \left(s + 1\right)$ operations (regression models and correlations), where $s$ denotes the number of selected features. 

In statistics, consistent model selection translates into selecting the correct features with probability tending to $1$ as the sample size tends to infinity. Zhang (2009) \cite{zhang2009} discussed the necessary and sufficient conditions for a greedy least squares regression algorithm (such as OMP) to select features consistently. These conditions match the necessary and sufficient conditions for LASSO mentioned by Zhao and Yu (2006) \cite{zhao2006}. The main research question however, is which algorithm performs better in practice for typical low-sample, high-dimensional omics datasets. In addition, whether the performance depends on the outcome type. In prior work, it has been noticed that LASSO tends to choose considerably more features than necessary, leading to an abundance of falsely selected features (false positives) \cite{bertsimas2016, su2017}. Of course, one can force LASSO to select fewer features by increasing the value of the regularization hyper-parameter, so the previous statement holds for the standard way to tuning LASSO. In contrast, OMP and other greedy algorithms, such as FBED, tend to select fewer features and lead to more parsimonious models \cite{borboudakis2019}.

\section{The  \normalfont \textbf{g\small{OMP ALGORITHM}}}
The generalised OMP (gOMP) algorithm, presented in Algorithm \ref{gomp}, generalises OMP towards the following directions:

\begin{itemize}
\item[a)] The OMP algorithm has been defined for univariate/multivariate continuous and binary outcomes using linear and logistic regression models, but any model (learning algorithm) could be used, denoted as the function $f$ in Lines \ref{line:modelinitial} and \ref{line:modelupdate}. Function $f$ accepts the data matrix $X$ and the outcome $y$ and outputs a model instance $M$. The latter is again a function that accepts a data matrix $X$ and outputs predictions $y$. For multi-class problems, $y$ contains a prediction for each class and sample, and thus it is a matrix. The main restrictions for $f$ are that it has to be relatively efficient to train $p$ times (the number of selected features), and that its degrees of freedom and the data likelihood have to be available for some statistical-based termination criteria to be used, discussed below. For example, for continuous outcomes, linear, quantile \cite{koenker2001}, and MM \cite{yohai1987} regression models could be used, while with count data, Poisson, quasi Poisson and negative binomial regression models \cite{greene2003} can be used.

\item[b)] In the established OMP algorithm, the stopping criterion is for the error between the observed and the reconstructed signal to be below a threshold value $\epsilon$ (see Algorithm \ref{omp}). Our proposed modification is to use a statistical-based criterion instead, expressed by function {\em Stopping}. The latter accepts the model before and after the next feature selection, and the data. Examples of possible instantiations of this function is to return the $p$-value of the hypothesis that the likelihood of the model is improved with the inclusion of each new feature; this requires an F-test or a likelihood ratio test to be computed between two nested models. The stopping criterion is in Line \ref{line:stopping} of Algorithm \ref{gomp}.

\item[c)] gOMP further generalises the type of residuals to be extracted from each regression model, \textit{denoted by $Resid$ in lines \ref{line:resid1} and \ref{line:resid2} of Algorithm \ref{gomp}}. {\em Resid} accepts the current model $M$ and the data. For generalised linear models for example, the raw, deviance and Pearson residuals can be extracted, while for survival regression, the list includes deviance, response and martingale residuals \cite{therneau1990}. It is worth noting that all types of residuals are continuous. For multi-class problems, there is a residual for each sample but also for each class, and hence, residuals $r$ form a matrix instead of a vector.

\item[d)] The original OMP algorithm selects the candidate feature that maximizes the absolute linear correlation with the current residual vector. This is generalized to {\em any type of pairwise association} denoted by the function $\mathit{Assoc}$ in line \ref{line:assoc} of Algorithm \ref{gomp}. {\em Assoc} accepts the residuals and the values of a feature $X_i$. It could return an absolute measure of association or a $p$-value (see discussion below). When both the residuals and the features are continuous the linear Pearson correlation or the Spearman correlation coefficient can be employed. Other correlation options include the distance correlation \cite{szekely2007} that also captures non-linear relationships but is computationally expensive. To check correlation of the residuals with a discrete feature, we propose the use of Analysis of Variance (ANOVA) \cite{montgomery2017}. For multinomial or multivariate resposes that produce multivariate residuals the use of Multivariate ANOVA (MANOVA) \cite{mardia1979} is suggested.

\item[e)] Associations between different types of features (mixed data) may not be in the same scale. Instead of comparing association values, we propose to convert all associations to $p$-values that are on the same scale, by testing the hypothesis that the association is zero. The residual-based heuristic, then selects the feature with the smallest such $p$-value. Converting to $p$-values automatically adjusts for different degrees of freedom, for example discrete features with different number of values. It is important to note some practical considerations. First, the $p$-values of these tests need to be calibrated. This is typically possible for parametric tests. Some non-linear metrics of association require permutations to produce calibrated $p$-values, which is computationally expensive. When the sample size is relatively high, in the first iterations several $p$-values will be smaller than the machine epsilon. They will numerically be computed as zero. The algorithm is then choosing blindly among these features. It is important instead for one to directly compute the logarithm of the $p$-value using algorithms such as \cite{tsamardinos2019}.
\end{itemize}

\makeatletter
\def\BState{\State\hskip-\ALG@thistlm}
\makeatother

\begin{algorithm}[H]
\centering
\begin{algorithmic}[1]
\BState \textbf{Input}: \text{Outcome values $y$, dataset $X$}
\BState {\bf Hyper-Parameters}: Functions $f, \mathit{Resid}, \mathit{Assoc}, \mathit{Stopping}$ 
\BState \textbf{Output}: \text{A subset $\mathcal{S}\subseteq \mathcal{X}$ of selected features.} 
\State $\mathcal{S}\leftarrow \emptyset$
\State Initialize current model $M \leftarrow f(y, X_\mathcal{S})$ \label{line:modelinitial}
\State Initialize previous model $M' \leftarrow \emptyset$
\State Initialize residuals ${r}\leftarrow \mathit{Resid(y, X_\mathcal{S}, M)}$ \label{line:resid1}
\While{$\mathit{Stopping}(M, M', y, X)$}  \label{line:stopping}
\State $X_* \leftarrow \argmax_{i \in \mathcal{X} \setminus \mathcal{S}} \mathit{Assoc}(r, X_i)$  \label{line:assoc}
\State $\mathcal{S} \leftarrow \mathcal{S}\cup \{X_*\}$
\State $M' \leftarrow M$
\State $M \leftarrow f(y, X_\mathcal{S})$  \label{line:modelupdate}
\State $r \leftarrow \mathit{Resid}(y, X_\mathcal{S}, M)$ \label{line:resid2}
\EndWhile \\
\Return $\mathcal{S}$
\end{algorithmic}
\caption{The generalised OMP algorithm}\label{gomp}
\end{algorithm}

gOMP can easily handle most types of outcomes, even less frequently used types of outcomes. LASSO, on the contrary is heavily dependant upon the outcome variable. Each type of outcome requires a different approach and the development of new, possibly highly algebraically tedious, algorithmic procedures. In the case of survival outcomes, gOMP can easily employ a Weibull, log-logistic, or log-normal regression model, whereas LASSO is designed for Cox and Weibull regression only. With circular data, gOMP can employ a projected normal regression model \cite{presnell1998}, whereas LASSO has not been developed yet for such data. With compositional data, gOMP can employ the Zero Adjusted Dirichlet regression model \cite{tsagris2018} and easily bypass the problem of zero values, while LASSO requires the development of the appropriate algorithm because the log-ratio transformation is not applicable. 

gOMP has been implemented and is publicly available in the R package \textit{MXM} \cite{tsagris2019}. Numerous types of response variables have been covered, such as univariate and multivariate continuous, left censored continuous, binary, nominal, ordinal, percentages, time-to-event, (strictly) positive valued and counts. For these response variable types various regression models can be employed, such as linear regression, MM regression, median regression, Tobit regression, (quasi) logistic regression, multinomial regression, ordinal regression, beta and quasi logistic regression, Cox and Weibull regression, (quasi) Poisson and negative binomial regression.  

\section{Simulation studies}
Before moving to the empirical evaluation of gOMP we will show some interesting results of a small scale simulation study using binary (case-control) and continuous outcomes. For both types of outcome, we generated $50,000$ features from a normal distribution and ranged the sample sizes from $100$ to $1000$ increasing by $100$ each time. The evaluation criteria were the true positive rate (TPR); the percentage of relevant features selected and the false discovery rate (FDR); the percentage of falsely selected features. Further, we compared the performance of the final predictive model produced by the features selected by each algorithm. 

We randomly chose $10$ features to linearly produce the outcome and then added Gaussian noise. Specifically for the continuous outcome we used a high signal to noise ratio, equal to $32.5$. This is practically a noiseless case and with a relatively large sample size there should be no falsely selected features. 

We used the R package \textit{MXM} \cite{lagani2017, tsagris2019} for the gOMP algorithm, and \textit{glmnet} \cite{friedman2010} for LASSO. All experiments were performed on an Intel Core i5-4690K CPU @3.50GHz, 32GB RAM desktop computer. 

\subsection{Cross-validation pipeline} \label{bbc1}
We conducted a fully-automated machine learning pipeline for assessing the performance of each FS method, that is a 10-fold Cross-Validation (CV). The data are randomly split into 10 folds and a random fold acts as a test set, whereas the remaining folds compose the training set upon which gOMP and LASSO (using a range of algorithm-dependent hyper-parameters) are applied. The predictive performance of the regression models produced on the train set, by each algorithm is estimated on the test set. This process is repeated for each fold and the results were aggregated. For the binary (case-control) outcome scenario stratified random sampling ensured that the ratio of the number of cases to the number of controls was kept nearly the same within each of the 10 folds. 

\subsection{Hyper-parameters of gOMP and LASSO} \label{tuning}
Within the 10-fold CV procedure, tuning of hyper-parameters takes place. With the case-control outcome, we chose $10$ threshold values for gOMP, spanning from $\chi^2_{0.95, 1}=3.84$ (the $95\%$ upper quantile of the $\chi^2$ distribution with $1$ degree of freedom), up to $\chi^2_{0.99, 1}$. This represents a range of different critical values corresponding to different significance levels, e.g. $(\chi^2_{0.95, 1},\ldots, \chi^2_{0.99, 1})$\footnote{The 1 degree of freedom is justified from the use of continuous features only.}. With the continuous outcome, we used the adjusted $R^2$ (this is related to the $F$-test) as the stopping criterion and chose $10$ equidistant stopping values, ranging from $0.05\%$ up to $0.5\%$. 

We assessed the predictive performance of the features selected by gOMP and LASSO on the same basis. gOMP performs FS only whereas LASSO simultaneously performs FS and regularization. Following \cite{borboudakis2019} we also performed LASSO regularization on the features selected by gOMP so as to compare the predictive performance of gOMP with LASSO on a more equal basis. We chose $10$ $\lambda$ values resulting in $100$ predictive models. The default number of values of $\lambda$ in LASSO is $100$, starting from 0 up to a data dependent maximum value, resulting again in $100$ predictive models. 

\subsection{Predictive performance metrics}
We used the AUC as the performance metric in the case-control outcome scenario. AUC represents the probability of correctly classifying a sample to the class it belongs to, thus takes values between 0 and 1, where 0.5 denotes random assignment. Unlike the accuracy metric (proportion of correctly classified samples), AUC is not affected by the distribution of the two classes (cases and controls). We used the MSE (Mean Squared Error) as the predictive performance metric with continuous outcomes. A zero value corresponds to perfect prediction, whereas random guessing occurs when the average is used.

When the samples are limited to at most a few hundreds, the final estimated predictive performance of the best model is optimistically biased (the performance of the chosen model is overestimated). To overcome this we applied the bootstrap-based bias correction method \cite{tsamardinos2018}, described below. 

After completion of the cross-validation, the predicted values produced by all predictive models across all folds are collected in an $n \times M$ matrix $P$, where $n$ is the number of samples and $M$ the number of trained models (corresponding to the different combinations of hyper-parameter values). We sample rows (predictions) of $P$ with replacement, termed in-sample values, whereas the non re-sampled rows are termed out-of-sample values. The performance of each trained model in the in-sample values is computed and the model with the optimal performance is selected. Its performance is computed in the out-of-sample values. The process of resampling and performance calculation is repeated $B$ times and the average performance (in the out-of-sample values) is returned. This estimated performance usually underestimates the true performance, but this negative bias is smaller than the optimistic uncorrected performance, the performance estimated from the cross-validation \cite{tsamardinos2018}.  
The only computational overhead is with the repetitive re-sampling and calculation of the predictive performance, i.e. no model is fitted nor trained. This makes the bootstrap-based bias correction method highly attractive compared to the computationally expensive nested cross-validation \cite{varma2006}, while bias correction is equally effective \cite{tsamardinos2018}.

\subsection{Number of selected features}
As mentioned in the Introduction, it is often the case that the primary goal of FS, especially in biological domains, is to identify the relevant features that assist on interpretability and knowledge discovery. Hence, we also examined the number of features each algorithm selected and related it to their predictive performance, aiming at produced models both parsimonious and highly accurate.

\subsection{Computational efficiency}
gOMP selects most features when its stopping criterion is low; $\chi^2_{0.95, 1}=3.84$ in our case. For LASSO we used the default number of 100 $\lambda$ values. For each sample size we computed the speed-up factor of gOMP to LASSO (computational cost of LASSO divided by the computational cost of gOMP) in order to provide their relevant, unit-free computational cost.

\subsection{Simulation studies results}
\begin{figure*}[!ht]
\begin{center}
\includegraphics[scale=0.38, trim = 30 0 0 0]{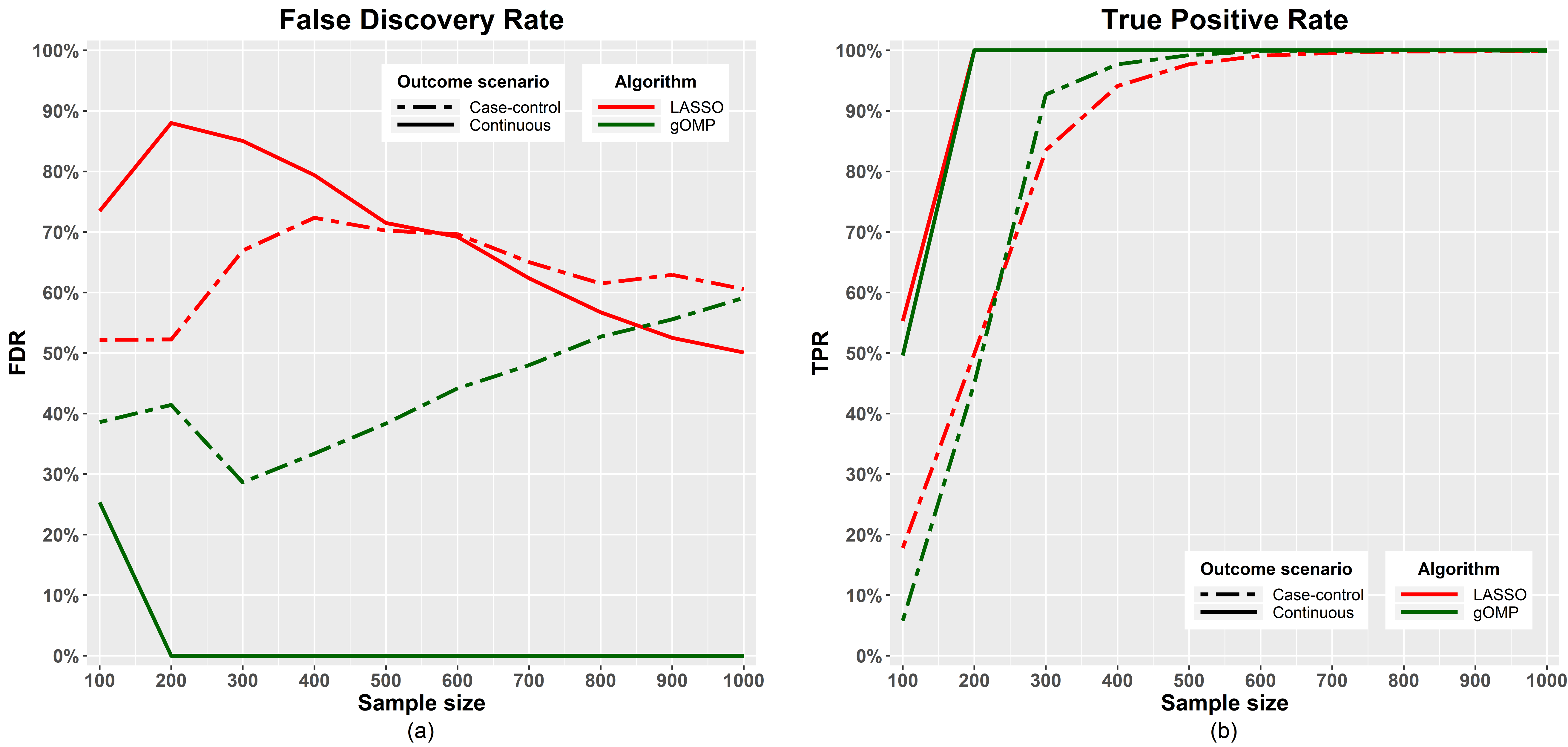} 
\caption{\textbf{The outcome is a linear function of a subset of the \underline{\textit{continuous}} features:} FDR (a) and TPR (b) for a range of sample sizes for gOMP and LASSO with binary and continuous outcome and $50,000$ features. (a) FDR ($y$-axis) for a range of sample sizes ($x$-axis). The algorithm ideally should have 0\% FDR and 100\% TPR. Hence, in (a) the points should lie close to 0\%, whereas in (b) the points should lie towards to 100\%. \label{fig_sim}}
\end{center}
\end{figure*}

\begin{figure*}[!ht]
\begin{center}
\includegraphics[scale=0.38, trim = 30 0 0 0]{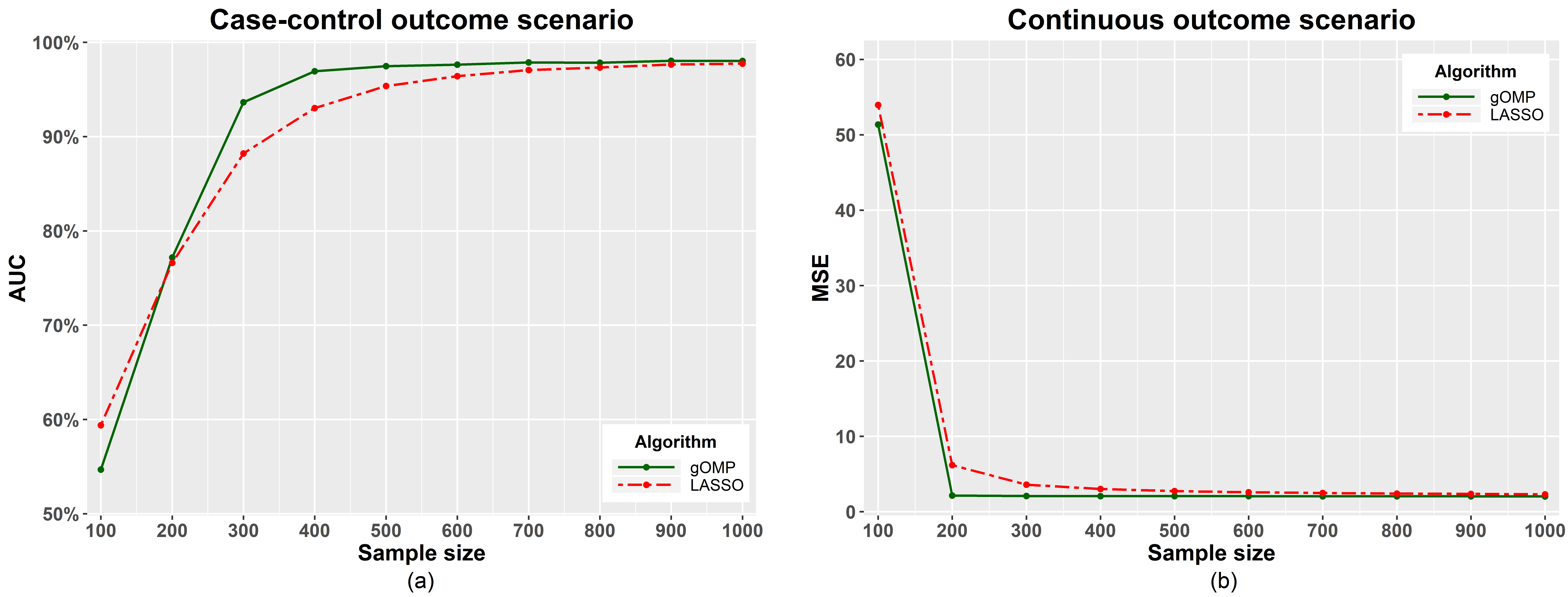} 
\caption{\textbf{The outcome is a linear function of a subset of the \underline{\textit{continuous}} features:} Performance of gOMP and LASSO for a range of sample sizes and $50,000$ features. (a) Case-control outcome: higher values of AUC are better. (b) Continuous outcome (signal to noise ratio is equal to $32.5$): lower values of MSE are better. \label{fig_perf_sim}}
\end{center}
\end{figure*}

Figure \ref{fig_sim} presents the results of TPR and FDR. Overall, the results show the superiority of gOMP in terms of TPR (higher is better) and FDR (lower is better). LASSO tends to select a higher number of irrelevant (not related to the outcome) features than gOMP. With large sample sizes, both gOMP and LASSO have a TPR equal to 100\%. However, gOMP achieves a 0\% FDR, whereas LASSO has an FDR equal to 50\%. Figure \ref{fig_perf_sim} shows the estimated performance of gOMP and LASSO for a range of sample sizes. For small to moderate samples gOMP outperforms LASSO, while for large sized samples, they produce comparable results. 

\section{Empirical evaluation using gene expression data}

Hastie et al. (2017) \cite{hastie2017} performed a simulation study comparing LASSO with other competing algorithms by using simulated datasets only. They concluded that LASSO was on par or outperformed two state-of-the-art FS algorithms, FSR \cite{weisberg1980} and best-subset-selection \cite{bertsimas2016}. Borboudakis and Tsamardinos \cite{borboudakis2019} compared FBED with LASSO using real high dimensional data from various fields, biology, text mining and medicine. Their results showed that LASSO produced predictive models with higher performance at the cost of being more complex. They concluded that there is no clear winner between LASSO and FBED and the choice depends solely on the goal of the analysis. 

The drawback of simulated data is that they do not necessarily portray the complex structure observed with real data. Following Borboudakis and Tsamardinos \cite{borboudakis2019} we also conducted extensive experiments focusing on real, publicly available, pre-processed gene expression data \cite{lakiotaki2018}. The gene expression data (GSE annotated) can be downloaded from \href{http://dataome.mensxmachina.org/}{BioDataome} \cite{lakiotaki2018}. For the case-control (and the continuous) outcome we searched for datasets with at least $100$ samples, while for the time-to-event outcome we included datasets with at least 70 samples, of which no more than $50\%$ were censored. 

\subsection{The three outcome scenarios}
We considered three different types of outcome variables: binary (case-control), continuous and (right censored\footnote{Censoring occurs when we have limited information about individual's survival time. The individual might have died of another cause, while the study was in progress or dropped out the study.}) survival times (time-to-event). 
\begin{itemize}
\item \textbf{Case-control outcome}\footnote{We focused on unmatched case-control gene expression data.}. The goal is to identify the minimal subset of genes that best discriminates among two classes. For gOMP we employed the logistic regression and similarly to Lozano et al. (2011) \cite{lozano2011}, we computed the raw residuals $e_i = y_i - \hat{y}_i$. In this scenario we investigated the performance of gOMP and LASSO using \textbf{53} gene expression datasets, with features (probesets) at the order of tens of thousands features ($17,000$, $22,000$, $33,000$, $45,000$ and $54,000$).

\item \textbf{Continuous outcome}. The next case scenario is when the variable of interest takes values in a continuum, the BMI index for instance. In this case, gOMP employed a linear regression model, computing again the raw residuals. We used the same \textbf{53} datasets as with the previous scenario. The feature (probeset) mostly correlated with the binary outcome played the role of the outcome.

\item \textbf{Time-to-event outcome}. The event of interest can be death (as in our case), a disease relapse, or in general any time-related event. The aim is to identify the subset of features, e.g. genes, mostly correlated with the survival time. Both gOMP and LASSO employed the Cox proportional hazards model. This is the only outcome for which we did not compute the raw residuals but the more appropriate martingale residuals \cite{therneau1990}, due to the nature of the problem. We note that this is the first time that a variant (or a generalisation) of OMP treats a survival outcome. For this outcome scenario we used \textbf{10} gene expression datasets\footnote{This very small number of datasets and the peculiarity of this type of outcome are the two main reasons we did not perform simulation studies for this type of outcome.}, whose dimensionality spans from a few thousands of features to tens of thousands of features. 
\end{itemize}

\subsection{Types of features}
gOMP, as mentioned in Algorithm \ref{gomp}, is able to treat not only continuous, but also categorical features via computation of p-values from the appropriate hypothesis test depending on the nature of the feature. We hence performed two types of experimental evaluation studies. The first one used the original datasets that contain continuous features only, whereas for the second type we introduced categorical features. Half of the features were randomly selected and were discretised according to their 33\%th and 66\%th quantile values. We used the R package \textit{MXM} \cite{lagani2017, tsagris2019} for the gOMP algorithm, and \textit{glmnet} \cite{friedman2010} for LASSO with continuous features. Group LASSO (GLASSO) that is available in the R package \textit{gglasso} \cite{yang2017} was used for the continuous and case-control outcomes. All experiments were performed on an Intel Core i5-4690K CPU @3.50GHz, 32GB RAM desktop computer.

\subsection{Cross-validation pipeline} \label{bbc}
We conducted a 10-fold cross-validation for the case-control and the linear outcomes, but an 8-fold cross-validation for the time-to-event outcome due to the relatively small number of samples. We also estimated the performance of LASSO when it selects, roughly, as many features as gOMP in order to compare them on a more equal basis. This was denoted by LASSO$^*$.

\subsection{Predictive performance metrics and number of selected features}
We used again the AUC as the performance metric in the case-control outcome scenario and the MSE for the continuous outcome scenario. For the time-to-event outcomes we used the concordance index (C-index) \cite{harrell1996} as the performance metric for model assessment. The C-index expresses the probability that, for a pair of randomly chosen samples, the sample with the highest risk prediction will be the first one to experience the event (e.g death). It measures the percentage of pairs of subjects correctly ordered by the model in terms of their expected survival time. A model ordering pairs at random (without use of any feature) is expected to have a C-index of 0.5, while perfect ranking would lead to a C-index of 1. When there are no censored values, the C-index is equivalent to the Area Under the Curve (AUC).

\subsection{Statistical evaluation of the predictive performance and the number of selected features}
We statistically evaluated the differences in predictive performance and number of selected features. For each evaluation criterion separately, we calculated the differences between gOMP \& LASSO. The average difference served as the observed test statistic. We then randomly permuted, $999$ times, the sign of the differences, and each time calculated the mean difference. The p-value was computed as the proportion of times the permuted test statistics exceeded the value of the observed test statistic. We repeated this process $1000$ times and reported the average p-value as a means of safer conclusions. 

\subsection{Computational efficiency}
We measured the computational efficiency of each algorithm during the $8$ or $10$-fold cross-validation. Similarly to the simulation studies, we computed the speed-up factor of gOMP to LASSO for a range of randomly selected subsets of features. The dimensionality of the features is different though. In order to make them comparable, we ranged the percentage of features from $2\%$ to $100\%$ with a step-size of $2\%$. 

\subsection{Empirical evaluation results}
\subsubsection{Predictive performance and number of selected \textit{continuous} features}
\begin{figure*}[!ht]
\begin{center}
\includegraphics[scale=0.50, trim = 30 0 0 0]{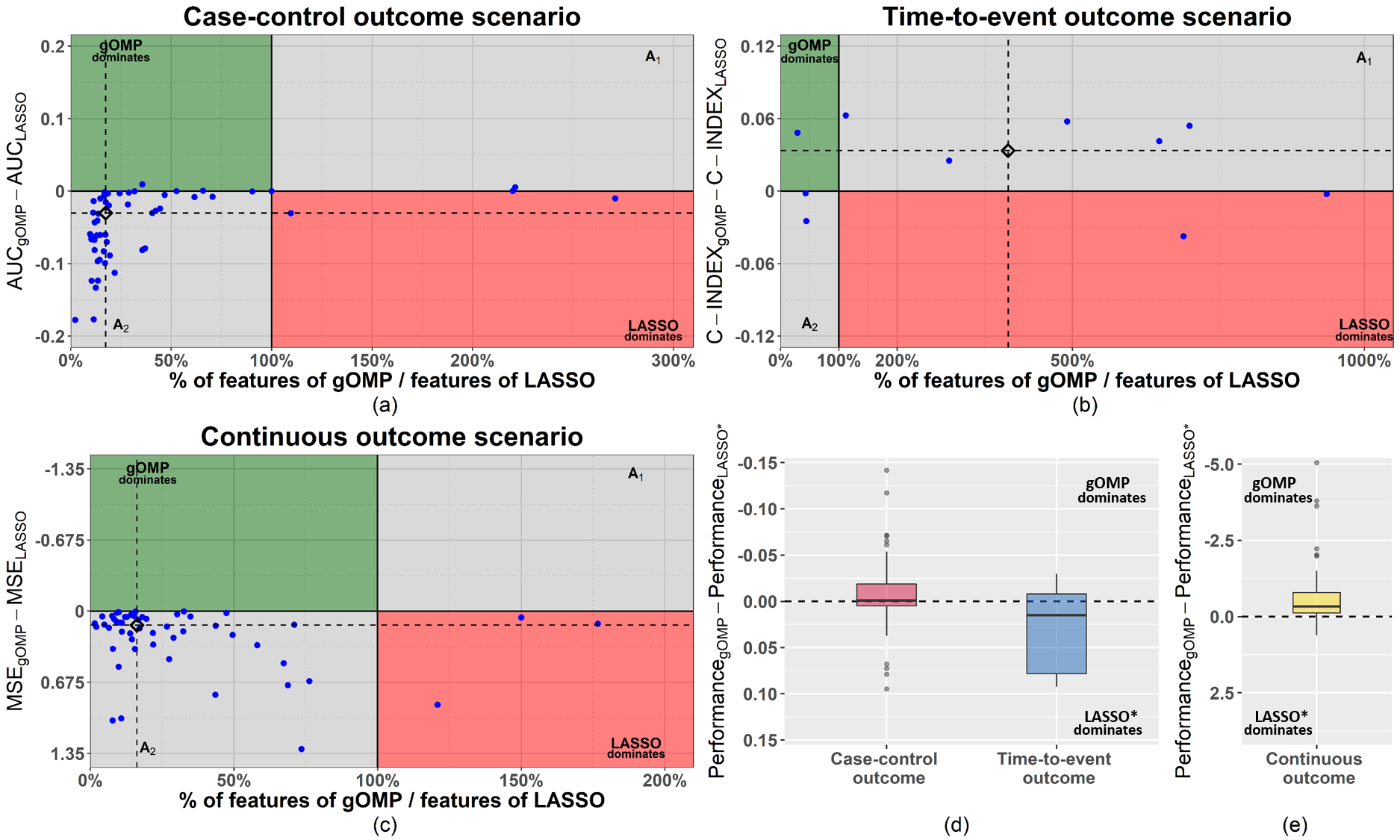} 
\caption{\textbf{\textit{Continuous} features:} \textbf{(a)-(c):} Predictive performance vs number of selected features for the different oucome scenarios. The $x$-axis represents the percentage-wise ratio of selected features between gOMP and LASSO. Values less than $100\%$ indicate that gOMP selected fewer features than LASSO. The $y$-axis represents the predictive performance difference between gOMP and LASSO, with positive values indicating that gOMP performs better than LASSO, except for the case of continuous outcomes as in \textbf{(c)}. In this case, negative values indicates that gOMP performs better than LASSO. gOMP outperforms LASSO in both performance and number of selected features in the top left quartile, shown in \textit{green}. LASSO outperforms gOMP in both performance metrics in the bottom right quartile (\textit{red} area). $A_1$: gOMP has better predictive performance than LASSO while selecting more features. $A_2$: LASSO has better predictive performance than gOMP while selecting more features. The rhombus sign corresponds to the median of the values in the axes. \textbf{(d)-(e):} Box plots of the differences in the predictive performance of gOMP and LASSO$^*$. For the case-control and the time-to-event outcome scenarios positive values indicate gOMP outperforms LASSO$^*$. For the continuous outcome scenario, gOMP outperforms LASSO$^*$ when the differences are negative. \label{fig1}}
\end{center}
\end{figure*}

Figures \ref{fig1}(a)-(c) present the difference in the predictive performance for each outcome scenario as a function of the number of selected features. Figure \ref{fig1}(d) presents box-plots of the difference in the predictive performance of gOMP and LASSO$^*$.

\begin{itemize}
\item \textbf{Case-control outcome scenario}. Figure \ref{fig1}(a) presents the results for the binary outcome scenario. LASSO achieved close or higher predictive performance than gOMP in most cases. LASSO though selected significantly more features (up to $10$ times more) than gOMP. The p-value for the mean difference in the AUC was equal to $0.001$, indicating that the differences in the predictive performance of gOMP and LASSO, based on all \textit{53} datasets, were statistically significant. LASSO selected statistically significantly more features than gOMP (p-value = $0.001$). The predictive performance of LASSO$^*$ was not significantly different than that of gOMP though (p-value = $0.200$). Figure \ref{fig1}(d) shows their differences in the AUC. 

\item \textbf{Time-to-event outcome scenario}. Figure \ref{fig1}(b) shows the relationship between the predictive performance (C-index) and the number of selected features. It is evident that in terms of predictive performance and number of selected features, gOMP outperformed LASSO, as it achieved better performance with a smaller set of features. The p-value for the mean difference in the C-index was equal to $0.089$ providing evidence that the predictive performance of gOMP was similar to that of LASSO. The p-value for the mean difference in the selected features was $0.922$. The predictive performance of LASSO$^*$ was similar to that of gOMP (p-value = $0.116$, Figure \ref{fig1}(d)). 

\item \textbf{Continuous outcome scenario}. Figure \ref{fig1}(c) presents the results for the continuous outcome scenario. LASSO achieved close or higher predictive performance than gOMP in most cases. The p-value for the mean difference in the PMSE was equal to $0.001$, indicating that the differences in the predictive performance of gOMP and LASSO, based on all \textit{53} datasets, were statistically significant. Once again, LASSO selected statistically significantly more features than gOMP (p-value = $0.001$). The predictive performance of LASSO$^*$ was significantly different than that of gOMP (p-value = $0.001$) though. This means, that on a more fair basis, when these two algorithms selected roughly equal number of features, gOMP performed significantly better than LASSO (see Figure \ref{fig1}(d)).
\end{itemize}

\subsubsection{Predictive performance and number of selected \textit{categorical} features}
\begin{figure*}[!ht]
\begin{center}
\includegraphics[scale=0.50, trim = 30 0 0 0]{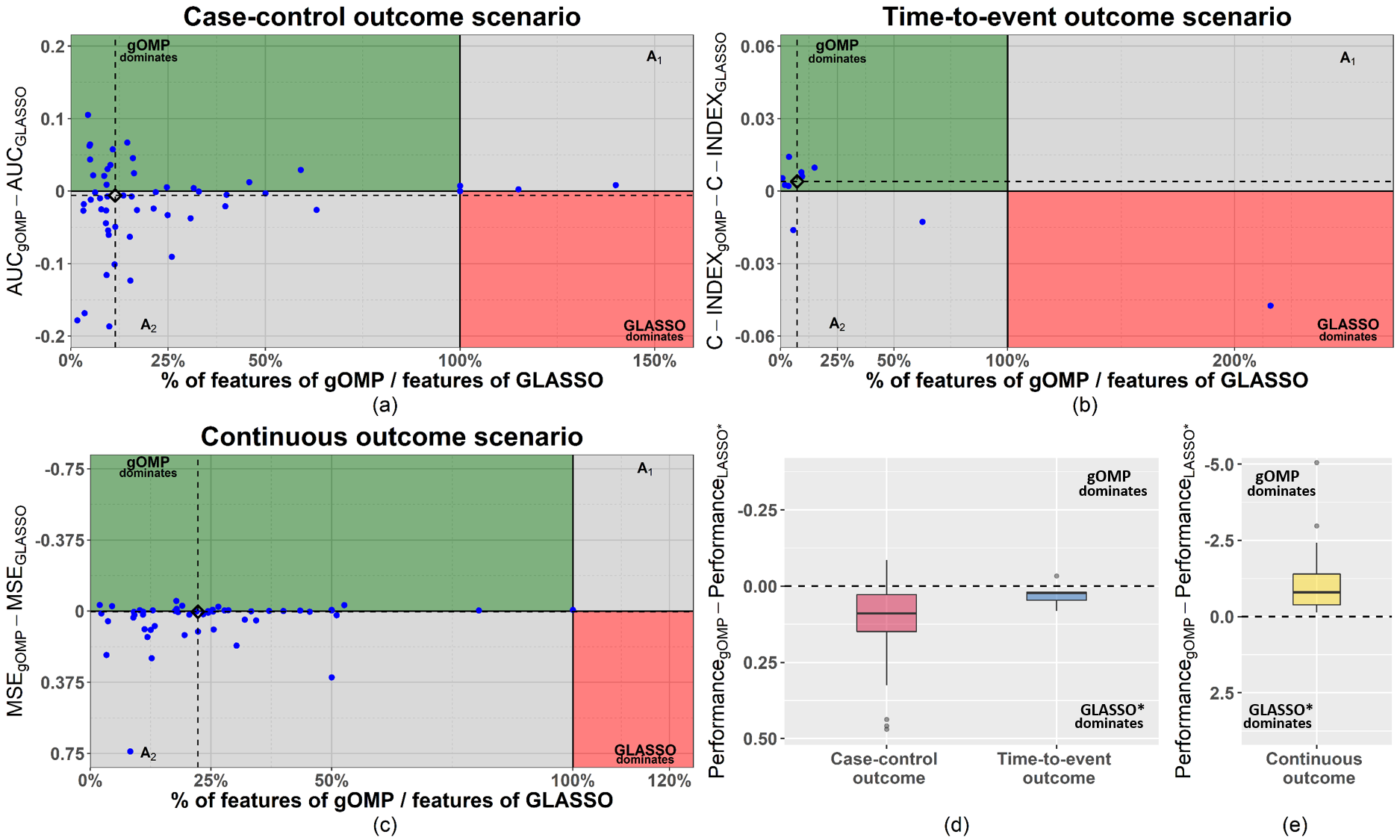} 
\caption{\textbf{\textit{Continuous and categorical} features:} \textbf{(a)-(c):} Predictive performance vs number of selected features for the different oucome scenarios. The $x$-axis represents the percentage-wise ratio of selected features between gOMP and GLASSO. Values less than $100\%$ indicate that gOMP selected fewer features than GLASSO. The $y$-axis represents the predictive performance difference between gOMP and GLASSO, with positive values indicating that gOMP performs better than GLASSO, except for the case of continuous outcomes as in \textbf{(c)}. In this case, negative values indicates that gOMP performs better than LASSO. gOMP outperforms LASSO in both performance and number of selected features in the top left quartile, shown in \textit{green}. GLASSO outperforms gOMP in both performance metrics in the bottom right quartile (\textit{red} area). $A_1$: gOMP has better predictive performance than GLASSO while selecting more features. $A_2$: LASSO has better predictive performance than gOMP while selecting more features. The rhombus sign corresponds to the median of the values in the axes. \textbf{(d)-(e):} Box plots of the differences in the predictive performance of gOMP and GLASSO$^*$. For the case-control and the time-to-event outcome scenarios positive values indicate gOMP outperforms LASSO$^*$. For the continuous outcome scenario, gOMP outperforms LASSO$^*$ when the differences are negative. \label{fig2}}
\end{center}
\end{figure*}

\begin{itemize}
\item \textbf{Case-control outcome scenario}. Figure \ref{fig2}(a) presents the results for the binary outcome scenario. gOMP was on par with GLASSO in terms of predictive performance (p-value = $0.982$), while selecting significantly less features (p-value = $0.001$). The predictive performance of gOMP was statistically significantly higher though than that of GLASSO$^*$ p-value=$0.001$. (see Figure \ref{fig2}(d)). Hence, on a more fair basis, when these two algorithms selected roughly equal number of features, gOMP performed significantly better than GLASSO.

\item \textbf{Time-to-event outcome scenario}. Figure \ref{fig2}(b) shows the relationship between the predictive performance (C-index) and the number of selected features. In terms of predictive performance and number of selected features, gOMP achieved similar performance to GLASSO (p-value=$0.558$) with a smaller set of features though (p-value=$0.001$). The predictive performance of gOMP was also similar to that of GLASSO$^*$ (p-value = $0.101$, Figure \ref{fig2}(d)). 

\item \textbf{Continuous outcome scenario}. Figure \ref{fig2}(c) presents the results for the continuous outcome scenario. gOMP achieved similar or higher predictive performance (smaller PMSE values) than GLASSO in most cases. The mean difference in the predictive performance (measured by PMSE) based on all \textbf{53} datasets was statistically significant (p-value=$0.001$), while GLASSO selected statistically significantly more features than gOMP (p-value = $0.001$). The predictive performance of GLASSO$^*$ was significantly different than that of gOMP (p-value = $0.001$, Figure \ref{fig2}(e)).
\end{itemize}

\subsection{Continuous and categorical features}
We also 

\subsubsection{Computational efficiency of the algorithms}
\begin{figure}[!ht]
\begin{center}
\includegraphics[scale=0.33, trim = 40 0 0 0]{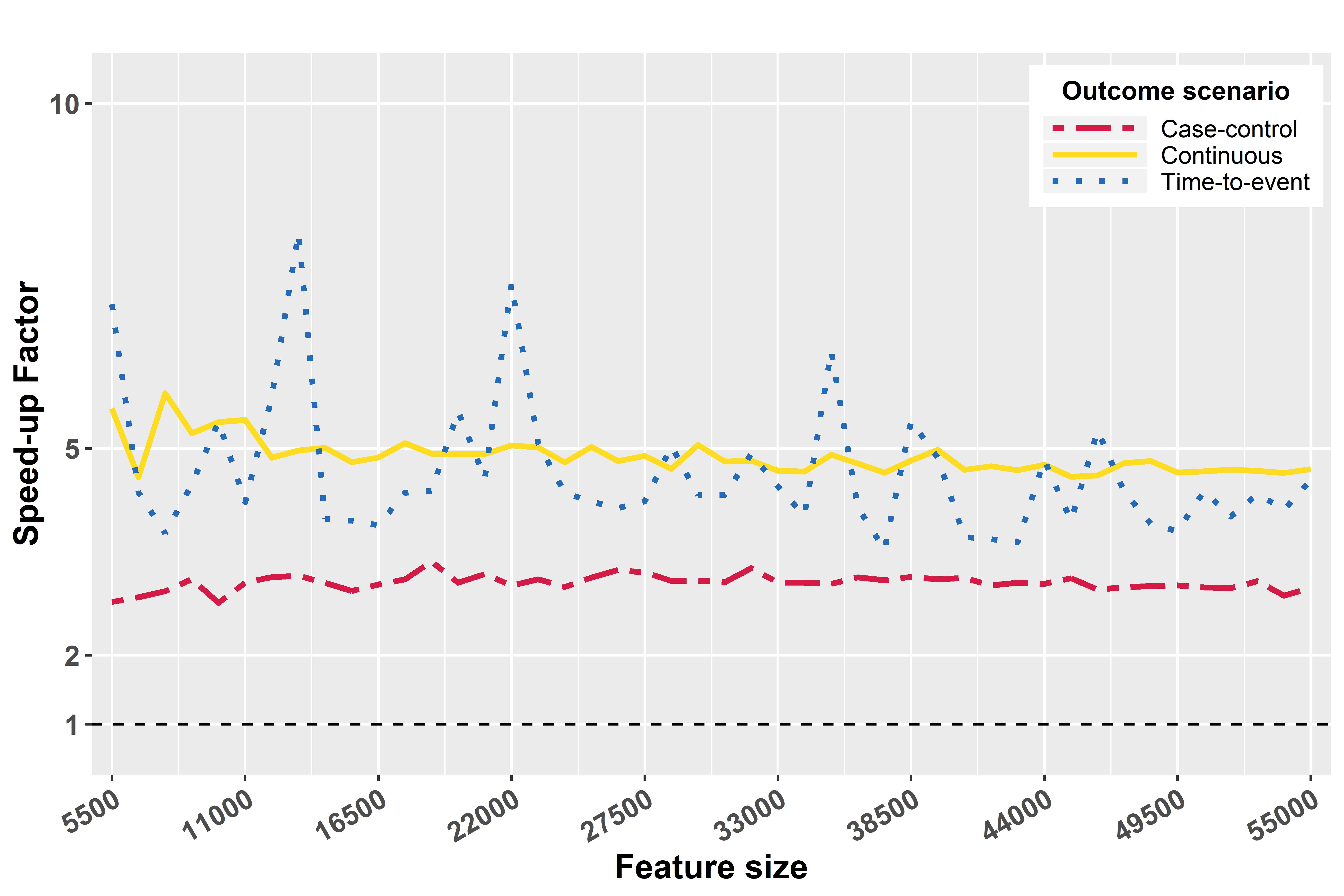} 
\caption{\textbf{Speed-up factors of gOMP versus LASSO ($y$-axis) for a range of \underline{\textit{continuous}} features ($x$-axis)}. \label{fig3}}
\end{center}
\end{figure}

\textbf{Continuous features:} The speed-up factors, for a range of subsets of features, are graphically presented in Figure \ref{fig3}. For the case-control outcome, gOMP is on average, $3$ times faster than LASSO, for the continuous outcome scenario, gOMP is on average $5$ whereas for the time-to-event outcome scenario, gOMP is on average $4.6$ times faster than LASSO. 

\begin{figure}[!ht]
\begin{center}
\includegraphics[scale=0.33, trim = 40 0 0 0]{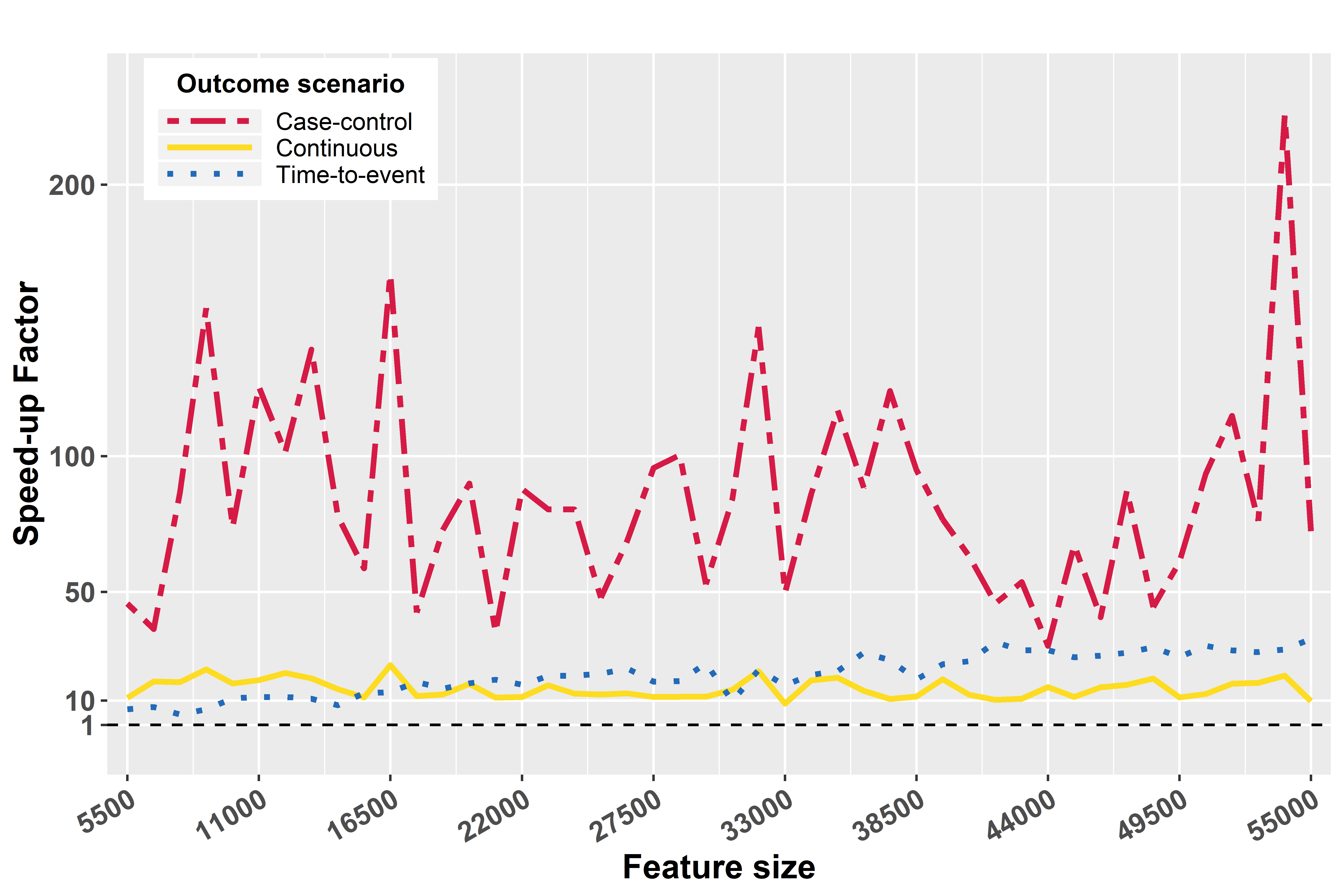} 
\caption{\textbf{Speed-up factors of gOMP versus GLASSO ($y$-axis) for a range of \underline{\textit{continuous and categorical}} features ($x$-axis)}. \label{fig4}}
\end{center}
\end{figure}

\textbf{Continuous and categorical features:} 
The speed-up factors, for a range of subsets of features, are graphically presented in Figure \ref{fig4}. For the case-control outcome, gOMP is on average, $83.8$ times faster than GLASSO, for the continuous outcome scenario, gOMP is on average $14.4$ whereas for the time-to-event outcome scenario, gOMP is on average $18.7$ times faster than GLASSO. 

\section{Discussion}
We introduced the generalised OMP feature selection algorithm and compared it to LASSO and GLASSO in the case of binary, continuous and survival outcomes using simulated data and real gene expression data. The key point of gOMP is that it can treat numerous types of outcome variables employing various regression models, whereas LASSO is not that flexible, i.e. each regression model requires its own appropriate mathematical formulation. LASSO, on the other hand, due to its computational efficiency and predictive performance has gained research attention in various data science fields and numerous extensions and generalisations have been proposed over the years. 

We showed that gOMP is a competing alternative to LASSO and GLASSO. However, this is not the first time a competing algorithm is suggested. Borboudakis and Tsamardinos \cite{borboudakis2019} showed that FBED achieves similar predictive performance to LASSO, at the cost of being computationally more expensive. To assess the quality of each FS algorithm we focused on three key elements, a) predictive performance, b) number of selected features and c) computational efficiency. In the simulated data gOMP outperformed LASSO in all aspects. In the real data our conclusions are as follows.

\begin{itemize}
\item {\textbf{Predictive performance and number of selected features:}}
With case-control outcome variables, LASSO outperformed gOMP. This was also the case for the continuous outcome scenario. With time-to-event outcome though, OMP leads to predictive models whose predictive performance is similar to LASSO. LASSO, tends to select more features than necessary, leading to more complex predictive models. This has two disadvantages: a) the models are more difficult (and computationally expensive) to tune, train and interpret, and b) if the primary goal is to identify the relevant features, then gOMP should be preferred, as LASSO will have selected more features which are irrelevant. When we compared gOMP to LASSO having selected similar number of features, we showed that gOMP was on par or outperformed LASSO and GLASSO.  

\item{\textbf{Computational efficiency:}}
gOMP was proved to be substantially more efficient than LASSO and GLASSO in all scenarios examined. We highlight that gOMP has been implemented in R, whereas LASSO and GLASSO are implemented in Fortran.
\end{itemize}

\section{Conclusions and future work}
Our simulation and empirical studies clearly pointed out that gOMP can be used succesfully instead of LASSO. LASSO is computationally efficient, with high predictive performance at the price of producing complex models by including many irrelevant features. gOMP, similarly to LASSO, tackles the FS problem from a geometrical standpoint, and fits a small number of regression models, thus also being computationally highly efficient. gOMP produces predictive models of similar predictive performance to LASSO but more parsimonious. gOMP's extra advantage is it can treat numerous types of outcomes and employ various regression models, many of which can be found in the R package \textit{MXM} \cite{lagani2017, tsagris2019}.

On a different direction, Ein-Dor et al. (2004) \cite{ein2004} demonstrated that multiple, equivalent prognostic signatures for breast cancer can be extracted just by analyzing the same dataset with a different partition in training and test set, showing the existence of several genes which are practically interchangeable in terms of predictive power. We provided evidence of this phenomenon since gOMP, and LASSO achieved, many times, similar performances by selecting different sets of features. SES (\cite{tsamardinos2012, lagani2017}) is one of the few FS algorithms that discovers statistically equivalent feature sets. More recently, Pantazis  et al. (2017) \cite{pantazis2017} proposed a, computational geometry based, solution for discovering equivalent sets of features using LASSO. Our ongoing research focuses on extending gOMP to identification of multiple sets of features that are statistically equivalent in terms of predictive performance. 

\section*{Acknowledgments}

We would like to acknowledge Stefanos Fafalios, Kleio Maria Verrou, Spyros Kotomatas and Tasos Tsourtis for providing constructive feedback. 

The research leading to these results has received funding from the European ResearchCouncil under the European Union’s Seventh Framework Programme (FP/2007-2013) /ERC Grant Agreement No. 617393.


\end{document}